\documentclass[letterpaper]{article} 
\usepackage{aaai25}  
\usepackage{times}  
\usepackage{helvet}  
\usepackage{courier}  
\usepackage[hyphens]{url}  
\usepackage{graphicx} 
\usepackage{natbib}  
\usepackage{caption} 
\usepackage{amssymb}
\usepackage{amsmath}
\usepackage{multirow}
\usepackage{graphicx}
\usepackage{makecell}
\usepackage{algorithm}
\usepackage{algorithmic}

\usepackage{newfloat}
\usepackage{listings}
\DeclareCaptionStyle{ruled}{labelfont=normalfont,labelsep=colon,strut=off} 
\floatstyle{ruled}
\newfloat{listing}{tb}{lst}{}
\floatname{listing}{Listing}
\title{PC-BEV: An Efficient  Polar-Cartesian BEV Fusion Framework for \\ LiDAR Semantic Segmentation}
\author{
    Shoumeng Qiu\textsuperscript{\rm 1}, 
    Xinrun Li\textsuperscript{\rm 2}, 
    Xiangyang Xue\textsuperscript{\rm 1},
    Jian Pu\textsuperscript{\rm 1}\thanks{Corresponding author}
}
\affiliations{
    \textsuperscript{\rm 1} Fudan University, Shanghai, China \\
    \textsuperscript{\rm 2} Bosch Corporate Research, China \\

    smqiu21@m.fudan.edu.cn, \{xyxue,jianpu\}@fudan.edu.cn, xinrun.li2@cn.bosch.com

}

\usepackage{bibentry}

\begin{document}

\makeatletter
\newcommand*\bigcdot{\mathpalette\bigcdot@{.7}}
\newcommand*\bigcdot@[2]{\mathbin{\vcenter{\hbox{\scalebox{#2}{$\m@th#1\bullet$}}}}}
\makeatother

\maketitle

\begin{abstract}

Although multiview fusion has demonstrated potential in LiDAR segmentation, its dependence on computationally intensive point-based interactions, arising from the lack of fixed correspondences between views such as range view and Bird's-Eye View (BEV), hinders its practical deployment. This paper challenges the prevailing notion that multiview fusion is essential for achieving high performance. We demonstrate that significant gains can be realized by directly fusing Polar and Cartesian partitioning strategies within the BEV space. Our proposed BEV-only segmentation model leverages the inherent fixed grid correspondences between these partitioning schemes, enabling a fusion process that is orders of magnitude faster (170$\times$ speedup) than conventional point-based methods. Furthermore, our approach facilitates dense feature fusion, preserving richer contextual information compared to sparse point-based alternatives. To enhance scene understanding while maintaining inference efficiency, we also introduce a hybrid Transformer-CNN architecture. Extensive evaluation on the SemanticKITTI and nuScenes datasets provides compelling evidence that our method outperforms previous multiview fusion approaches in terms of both performance and inference speed, highlighting the potential of BEV-based fusion for LiDAR segmentation. Code is available at \url{https://github.com/skyshoumeng/PC-BEV.}

\end{abstract}

%

\begin{figure}[t]
\centering
\includegraphics[width=1.\linewidth]{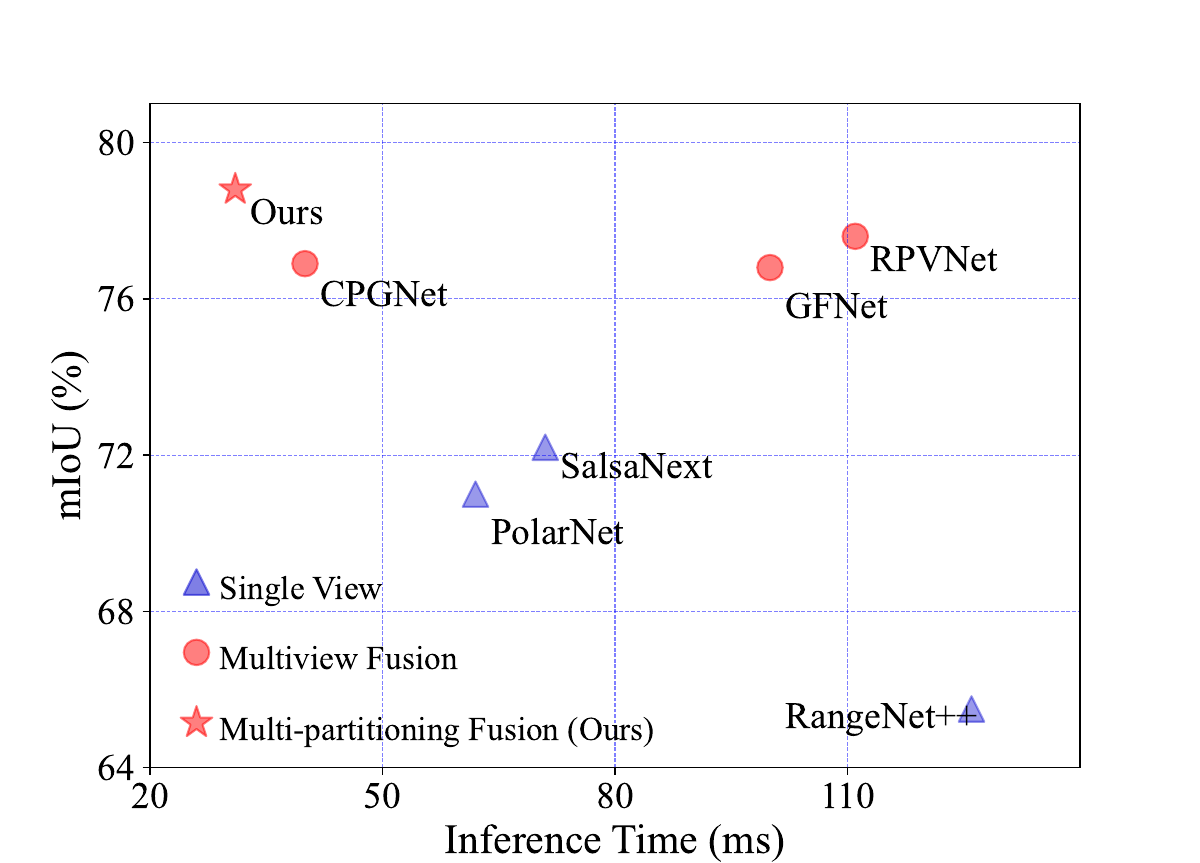}
\caption{Comparison with other projection-based methods, the results demonstrates the advantages of our method over others in terms of both performance and speed. Experiments are conducted on the nuScenes validation set.}
\label{fig:overview}
\end{figure}

\section{Introduction}

LiDAR point cloud segmentation is an essential perception task in autonomous driving systems \cite{li2020deep, gao2021we, feng2020deep}, aiming to provide fine-grained semantic understanding of the surrounding environment at the point level. Existing methods predominantly fall into three categories: point-based \cite{qi2017pointnet, tatarchenko2018tangent, cheng2020cascaded, thomas2019kpconv}, voxel-based \cite{choy20194d, tang2020searching, zhu2021cylindrical, ye2021drinet}, and projection-based approaches \cite{milioto2019rangenet++, alonso20203d, zhao2021fidnet, cheng2022cenet, ando2023rangevit}. Among these, projection-based methods, which leverage the efficiency of 2D convolutional neural networks (CNNs) on projected point clouds, have gained popularity due to their real-time inference capabilities, a critical requirement for autonomous vehicles. However, the inherent information loss during the 3D-to-2D projection process often limits their performance compared to computationally intensive voxel-based methods \cite{cheng20212}.

Multiview fusion has emerged as a promising solution to bridge this performance gap by exploiting the complementary information captured by different projection techniques \cite{liong2020amvnet, xu2021rpvnet}. Recent multiview fusion methods, such as AMVNet \cite{liong2020amvnet}, GFNet \cite{qiu2022gfnet}, and CPGNet \cite{li2022cpgnet}, employ point-based feature interactions across views to enhance representation learning. However, these methods suffer from inefficiencies stemming from the lack of fixed correspondences between views, necessitating computationally expensive grid sampling and scatter operations that impede real-time performance \cite{siam2018comparative, liu2023real, sanchez2023domain}. Furthermore, feature fusion is restricted to areas where points are present, potentially overlooking valuable contextual information in the surrounding regions.

To address these limitations, we propose a novel multi-partition feature fusion framework that operates solely within the Bird's-Eye View (BEV) space, leveraging the fixed correspondence between two commonly used Polar and Cartesian partitioning schemes. Our approach is motivated by the observation that Polar partitioning in BEV shares similarities with spherical coordinate partitioning in range view, and the experiments that the performance of different partitioning methods exhibits complementarity.

To facilitate feature fusion between the Polar and Cartesian branches, we introduce an efficient and effective remap-based fusion method. Leveraging the fixed coordinate correspondence inherent to Polar and Cartesian spatial partitioning within the same BEV space, our method precomputes correspondence parameters, enabling efficient feature fusion through carefully designed remapping operations. This approach is notably $170\times$ faster than previous point-based feature interaction methods. Furthermore, our feature fusion operates on all positions within the BEV space, achieving dense fusion and preserving more valuable contextual information, in contrast to sparse fusion in previous point-based methods, which is limited to regions where points exist. We also propose a hybrid Transformer-CNN architecture for BEV feature extraction. Self-attention within the Transformer block capture global scene information, followed by a lightweight U-net like CNN \cite{ronneberger2015u} for detailed feature extraction. Experimental results demonstrate that this architecture enhances model performance while maintaining real-time inference capabilities. 

The main contributions of our work can be summarized as follows:

\begin{itemize}
    \item We introduce the Polar-Cartesian BEV Fusion Network for LiDAR segmentation. In contrast to previous methods that fuse features from multiple views, our approach innovatively fuses features derived from distinct spatial partitioning strategies within a unified BEV representation. We also propose an effective hybrid network architecture for comprehensive BEV feature extraction. 

    \item We present an efficient and effective remap-based feature fusion method for integrating information from the two BEV representations. This method significantly outperforms previous point-based interaction approaches, achieving a 170$\times$ speedup while preserving richer contextual information during the fusion process.

    \item We conduct extensive experiments on the SemanticKITTI and nuScenes datasets, demonstrating that our method achieves state-of-the-art performance with faster inference speeds. Additionally, we provide detailed ablation studies and in-depth discussions of each module within our proposed framework. 
\end{itemize}

\begin{figure*}[t]
\centering
\includegraphics[width=.95\linewidth]{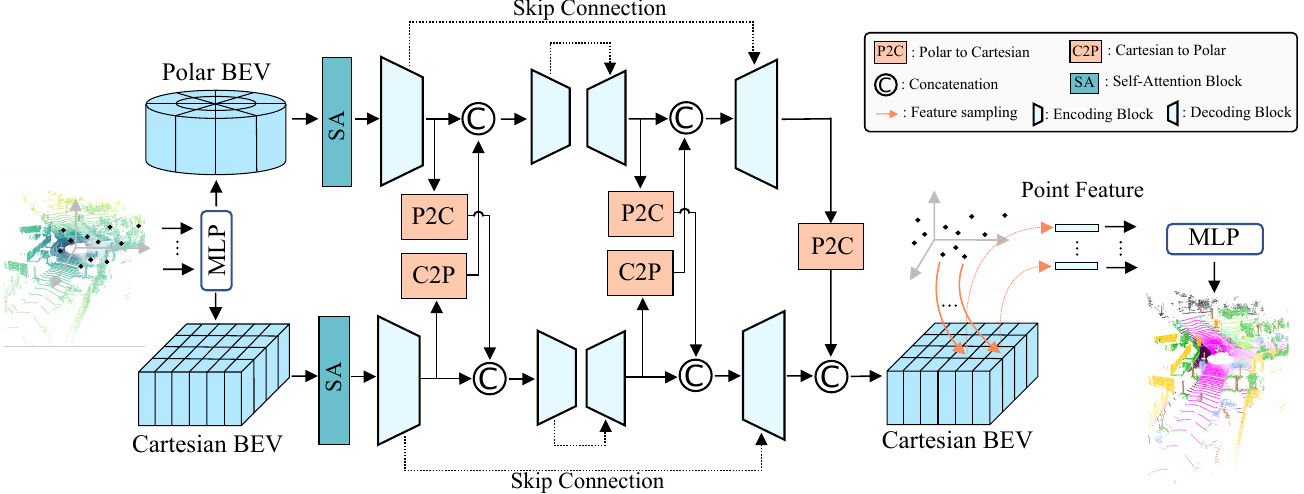}
\caption{The pipeline of our proposed Polar-Cartesian BEV fusion framework for 3D point cloud semantic segmentation task. Given a scan of point cloud, it first projected to a polar and a cartesian BEV pseudo-images as input to the Transformer-CNN Mixture architecture feature extraction network. Then the features between the two branches interact with each other bidirectional through the proposed effective PolarToCart (P2C) and CartToPolar (C2P) modules. Finally, we adopt the grid sampling operation to obtain the point-wise features from the concatenated features, and the sampled features are fed into a simple MLP block to obtain the final semantic predictions.}
\label{fig:overview}
\end{figure*}

\section{Related works}

\subsection{Projection-based Methods}

Projection-based methods generally provide faster inference and competitive performance compared to point-based and voxel-based methods. They can be further divided into two different categories: Range view projection (RV) and Bird's-Eye-View projection (BEV). For the range-based methods: RangeNet++ \cite{milioto2019rangenet++} proposed to exploit the potential of range images and 2D convolutions, and a GPU-accelerated post-processing K-Nearest-Neighbor (KNN) approach is further proposed to recover consistent semantic information during inference for the entire LiDAR scans. KPRNet \cite{kochanov2020kprnet} improved the convolutional neural network architecture for the feature extraction of the range image, and the commonly used post-processing techniques such as KNN were replaced with KPConv \cite{thomas2019kpconv}, which is a learnable pointwise component and allows for more accurate semantic class prediction. RangeViT \cite{ando2023rangevit} introduce ViTs to range view with pre-trained ViTs, substituting a tailored convolutional stem for the classical linear embedding layer, and pixel-wise predictions refinement module. RangeFormer \cite{kong2023rethinking} proposed introduces a comprehensive framework featuring innovative network architecture, data augmentation, and post-processing strategies that enhance the handling of LiDAR point clouds from the range view. It also proposes a Scalable Training from Range View (STR) strategy for effective training. For BEV-based approaches, PolarNet \cite{zhang2020polarnet} proposed to use the polar BEV representation to balance the spatial distribution of points in the coordinate system, and a ring convolution operation was also developed that was more suitable for the polar BEV representation. Panoptic-PolarNet \cite{zhou2021panoptic} was proposed based on PolarNet, which is a proposal-free LiDAR panoptic segmentation network that can cluster instances on top of the semantic segmentation efficiently. 

\subsection{Multiview Fusion-based Methods}
Point-based, voxel-based, and projection-based methods have different advantages while also suffering from their own shortcomings in the semantic segmentation task \cite{xu2021rpvnet}. Therefore, it is reasonable to fuse information from different views to enhance segmentation performance. Tornado-net \cite{gerdzhev2021tornado} proposed a new deep convolutional neural network for semantic segmentation that leveraged multiview (BEV and Range) feature extraction and encoder-decoder ResNet architecture with a proposed diamond context block. AMVNet \cite{liong2020amvnet} proposed an assertion-based multiview(BEV and Range) fusion network for LiDAR semantic segmentation, where the features of individual views were fused later with an assertion-guided sampling strategy. RPVNet \cite{xu2021rpvnet} devised a deep fusion framework with multiple and mutual information interactions among three (Range, Point, and Voxel) different views to make feature fusion more effective and flexible. GFNet \cite{qiu2022gfnet} introduced a geometric flow network to better explore the geometric correspondence between two different views (BEV and Range) in an align-before-fuse manner. CPGNet \cite{li2022cpgnet} proposed a Cascade Point-Grid Fusion Network equipped with a Point-Grid fusion block for the feature interaction (BEV and Range) and a transformation consistency loss narrows the gap between the single-time model inference with different transformed point clouds as input.

\section{Method}

\subsection{Polar-Cartesian BEV Fusion  Framework}

The overview of our proposed Polar-Cartesian BEV Fusion framework for LiDAR semantic segmentation is shown in Figure \ref{fig:overview}. It includes two branches: Cartesian branch and Polar branch. Given a point cloud $P=\{p_0, p_1, …, p_{N-1}\}$, which consists of $N$ LiDAR points $p_i = \{x_i, y_i, z_i, r_i\}$, where $\{x_i, y_i, z_i\}$ are the Cartesian coordinates relative to the scanner and $r_i$ is the intensity of the returning laser beam. We apply two distinct partitioning strategies for the BEV projection: Cartesian and Polar. The point cloud is quantized only along the $x$ and $y$ axes for efficient 2D-based feature extraction. For the point cloud $P$, the points are first encoded by a simplified PointNet \cite{qi2017pointnet}, which consists of only fully connected layers, batch normalization, and ReLU layers. Subsequently, the extracted features are scattered back into the BEV space, represented as $F_{cart}^{in}$ and $F_{polar}^{in}$ respectively. We employ two networks with identical structures but different parameters to perform feature extraction. Features from these two branches undergo bidirectional interactions during the process, including feature alignment and fusion, and the final prediction results are also derived from the fusion of these two branches.

\begin{figure*}[t]
\centering
\includegraphics[width=.9\linewidth]{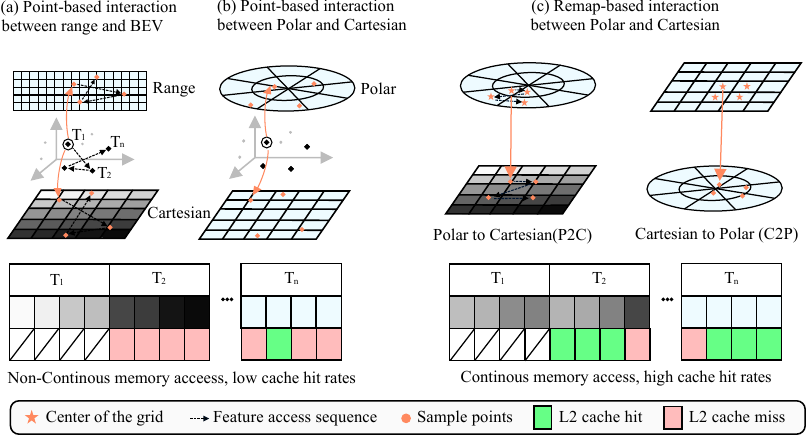}
\caption{Comparisons of the feature interaction operation processes between the previous point-based method and our proposed remap-based method across different settings. $\{\mathrm{T}_1, \mathrm{T}_2, \dots, \mathrm{T}_n\}$ denotes the CUDA kernel processing at different time steps and the corresponding cache states. Point-based method uses points as a bridge to facilitate feature interaction across different perspectives or spatial partitioning strategies, while our remap-based method relies on fixed corresponding in the same BEV space. For the point-based method, each point is treated as an individual for the point-level parallelism, and the fused features are ultimately scattered back to the original feature space. Our method, however, leverages advantage of spatial continuity during the remap process to reduce cache missing, enabling more efficient parallel processing, and eliminating the need for scatter back operations, significantly enhancing computational efficiency. }
\label{fig:comp_gs_rb}
\end{figure*}

The feature extraction network for each branch is our proposed Transformer-CNN Mixture architecture, which includes two standard transformer \cite{vaswani2017attention} blocks and a CNN network. The details of the architecture will be introduced in the \emph{Transformer-CNN Mixture Architecture} subsection. For the CNN model, We followed the model structure as described in \cite{li2022cpgnet}.

For the bidirectional feature interaction between the two branches, we first perform spatial alignment on the features from both branches. Specifically, suppose the features $F_{polar}$ and $F_{cart}$ come from the Polar and Cart branches, respectively, we use a Polar to Cartesian and Cartesian to Polar remapping operation to align the spatial features under the different partitioning strategies. 
\begin{equation}
\begin{aligned}
\hat{F}_{cart} = \operatorname{PolarToCart}(F_{polar}), 
\end{aligned}
\label{equ:wg}
\end{equation}
\begin{equation}
\begin{aligned}
\hat{F}_{polar} = \operatorname{CartToPolar}(F_{cart}),
\end{aligned}
\label{equ:wg}
\end{equation}
where $\operatorname{PolarToCart}(\cdot)$ and $\operatorname{CartToPolar}(\cdot)$ refer to the remapping from Polar space to Cartesian space and from Cartesian space to Polar space, respectively. The details of the remapping operation will be illustrated in the \emph{Feature Fusion through Remapping} subsection. For feature fusion, we adopt the commonly used concatenation operation \cite{park2017rdfnet,zhang2019multi,li2020enhanced}. For example, when fusing features from the Polar branch into Cartesian branch, we first concatenate the spatially transformed feature \(F'_{cart}\) with \(F_{cart}\). Then, we use a simple convolution operation to reduce the channel size of the feature to its original size. The fusion process can be expressed as:
\begin{equation}
\begin{aligned}
F^{fusion}_{cart} = \operatorname{MLP}_{fusion}(\operatorname{Concat}(\hat{F}_{cart}, F_{cart})), 
\end{aligned}
\label{equ:wg}
\end{equation}

For the final semantic prediction, since we aim to provide the semantic prediction for every point in the scene, so we need to acquire the features for class prediction in the projection space for each point. Given that we have features extracted from different branches, the common approach in previous methods involves retrieving the corresponding features from each branch for every point, typically through a grid sampling (GS) operation. Then the features sampled from different branches are fused. Finally, the fused features are used to obtain the final semantic prediction result. The previous point-based output fusion can be expressed as (here we assume use the concatenation operation for fusion):
\begin{equation}
\begin{aligned}
F_{out} =  \operatorname{Concat}(\operatorname{GS}(F^{out}_{cart}), \operatorname{GS}(F^{out}_{polar})).
\end{aligned}
\label{equ:wg}
\end{equation}
To further speed up the model inference, we use the remapping operation to align the features from one branch to another, which allows us to perform grid sampling just once on the remapped branch. In this paper, we choose to align the features extracted from the Polar branch to the Cartesian space, as we experimentally found that this has a slightly better performance than the reverse way. We concatenate the remapped Polar features with the Cartesian features, then employ grid sampling to obtain the features at the corresponding BEV positions for each point. Consequently, the final point-level feature output in our approach can be expressed as:
\begin{equation}
\begin{aligned}
F_{out} =  \operatorname{GS}(\operatorname{Concat}(F^{out}_{cart}, \operatorname{PolarToCart}(F^{out}_{polar})))
\end{aligned}
\label{equ:wg}
\end{equation}
where \(F_{out} \in \mathbb{R}^{N \times C_{out}}\). Finally, The fused features are fed into the final semantic classifier:
\begin{equation}
\begin{aligned}
pred = \operatorname{MLP}_{cls}(F_{out}) 
\end{aligned}
\label{equ:wg}
\end{equation}
\subsection{Feature Fusion through Remapping}
\label{FFtR}
Unlike Previous multiview fusion methods operate in different projection spaces with dynamic grid-to-grid correspondence because of the information loss in projection, our method benefits from a fixed positional correspondence between the two partitioning branches under the same BEV space, which provides us with an opportunity to improve the efficiency of the feature fusion process. Specifically, we employ a remap technique to align the features under the two different partitioning methods. Given that the grid correspondence is fixed between two branches, the remapping parameters can be precomputed for efficient feature fusion. Below, we provide detailed information of the remap operation, highlighting the advantages of remap-based interaction over point-based interaction. We take the remapping process from polar space to Cartesian space as an example, noting that the remapping from Cartesian to polar space adheres to the same principle.

\begin{figure}[t]
\centering
\includegraphics[width=.8\linewidth]{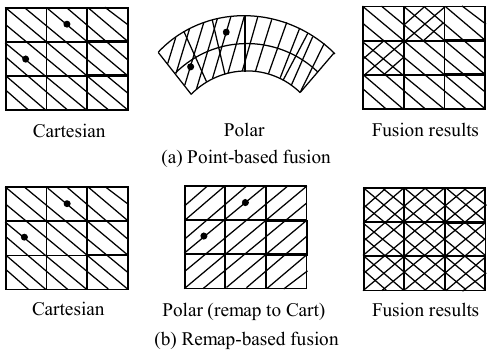}
\caption{Comparisons between point-based interaction results and our proposed remap-based interaction results, $\bigcdot$ denotes the LiDAR points.  The point-based method only fuses features where points exist, resulting in sparse fusion, while our method performs fusion across the entire space, resulting in dense fusion that incorporates more information.}
\label{fig:gs_rb_fusion}
\end{figure}

For each grid in the Cartesian branch, we denote the coordinates of the grid center as $\{Pos_{\{0,0\}}, Pos_{\{0,1\}}, Pos_{\{1,0\}}, ..., Pos_{\{H-1,W-1\}} \}$, where $Pos_{\{i,j\}} = \{p_i, p_j\}$. Next, we need to identify their corresponding coordinates in the Polar branch for feature fusion. To achieve this, we first compute the coordinates of grid center point $Pos_{\{i,j\}}$ in the real-world BEV space  $\{x_i, y_i\}$. Subsequently, we can easily calculate the coordinates of each point under the Polar branch adhering to the Polar partitioning mechanism: $ phi=\operatorname{arctan2}\{x_{i}, y_{j}\}, rho=\sqrt{x_{i}^2 + y_{j}^2}$. Up to now, the coordinate correspondence between the Cartesian and Polar branches is established, which is fixed so we can precompute before fusion. We could treat the grid center as a point and apply the previous point-based method for feature fusion; however, our experiments show that this approach is inefficient in practice.

For more efficient and effective feature fusion, we developed a remap-based feature fusion operation that significantly enhances the feature alignment speed between the two branches. The traditional point-based methods is slow mainly because the gird sampling operation and the scatter back operation. They treat each point individually for the point-level parallelism, resulting in a high cache missing rate in experiments. Unlike the point-based method, our remap-based operation considers the continuity of spatial positions, making the process more memory-access-friendly, and significantly speeding up computation. Figure \ref{fig:comp_gs_rb} presents a comparison of different feature fusion methods. It should be noted that not every grid in one branch has a corresponding area in another branch due to variations in space occupancy patterns. In practice, if a spatial location in one branch method is unavailable in the other, we simply apply zero padding to that location. More detailed efficiency analysis can be found in the supplementary materials.

Our remap-based fusion method offers additional advantages by incorporating more contextual information during the fusion process. As shown in Figure \ref{fig:gs_rb_fusion}, the point-based method only performs fusion in areas where points exist, discarding features where there are no points, a process we refer to as sparse fusion. In contrast, our remap-based method enables fusion throughout the entire BEV space, resulting in dense fusion that enriches features with more information from the other branch.

\begin{table*}[t]
  \centering
  \small
  \caption{Quantitative comparison of our Polar-Cartesian BEV Fusion Framework (PC-BEV). The results are reported in terms of the mIoU on the SemanticKITTI validation set. }
    \setlength{\tabcolsep}{1.mm}{
    \begin{tabular}{l|lllllllllllllllllll|l}
    \hline
    & & & & & & & & & & & & & & & & & & & &\\
    & & & & & & & & & & & & & & & & & & & &\\
    \multicolumn{1}{l|}{Methods}  & {\rotatebox{70}{car}} & {\rotatebox{70}{bicycle\hspace{-1cm}}} & {\rotatebox{70}{motorcycle\hspace{-1cm}}} & {\rotatebox{70}{truck\hspace{-1cm}}} & {\rotatebox{70}{other-vehicle\hspace{-1cm}}} & {\rotatebox{70}{person\hspace{-1cm}}} & {\rotatebox{70}{bicyclist\hspace{-1cm}}} & {\rotatebox{70}{motorcyclist\hspace{-1cm}}} & {\rotatebox{70}{road}} & \multicolumn{1}{l}{\rotatebox{70}{parking}} & {\rotatebox{70}{sidewalk}} & {\rotatebox{70}{other-ground\hspace{-1cm}}} & {\rotatebox{70}{building\hspace{-1cm}}} & {\rotatebox{70}{fence\hspace{-1cm}}} & {\rotatebox{70}{vegetation\hspace{-1cm}}} & {\rotatebox{70}{trunk\hspace{-1cm}}} & {\rotatebox{70}{terrain\hspace{-1cm}}} & {\rotatebox{70}{pole\hspace{-1cm}}} & {\rotatebox{70}{traffic-sign\hspace{-1cm}}} & \multicolumn{1}{l}  {mIoU} \\    \hline

    \multicolumn{1}{l|}{RangeNet++ } &  91.0 & 25.0 & 47.1 & 40.7 & 25.5 & 45.2 & 62.9 & 0.0 & 93.8 & 46.5 & 81.9 & 0.2 & 85.8 & 54.2 & 84.2 & 52.9 & 72.7 & 53.2 & 40.0 & 52.8 \\  
    
    
    

    \multicolumn{1}{l|}{PolarNet }  & 91.5 & 30.7 & 38.8 & 46.4 & 24.0 & 54.1 & 62.2 & 0.0 & 92.4 & 47.1 & 78.0 & 1.8 & 89.1 & 45.5 & 85.4 & 59.6 & 72.3 & 58.1 & 42.2 & 53.6 \\  

    \multicolumn{1}{l|}{GFNet } & 94.2 & 49.7 & 63.2 & 74.9 & 32.1 & 69.3 & 83.2 & 0.0 & 95.7 & 53.8 & 83.8 & 0.2 & 91.2 & 62.9 & 88.5 & 66.1 & 76.2 & 64.1 & 48.3 &  63.0\\

     \multicolumn{1}{l|}{AMV (BEV) } &  94.0 & 30.0 & 56.0 & 64.6 & 42.9 & 60.6 & 77.3 & 0.0 & 93.5 & 41.2 & 79.3 & 0.2 & 88.6 & 46.3 & 86.5 & 56.0 & 73.2 & 62.5 & 45.8 &  58.9\\

    \multicolumn{1}{l|}{AMV (RV) } &  90.4 & 31.9 & 57.6 & 79.8 & 45.7 & 61.9 & 64.9 & 0.0 & 95.3 & 48.9 & 81.8 & 0.8 & 85.3 & 59.7 & 84.1 & 58.8 & 69.9 & 53.4 & 44.7  &  59.6\\

    \multicolumn{1}{l|}{AMVNet } & 95.6 & 48.8 & 65.4 & 88.7 & 54.8 & 70.8 & 86.2 & 0.0 & 95.5 & 53.9 & 83.2 & 0.3 & 90.9 & 62.1 & 87.9 & 66.8 & 74.2 & 64.7 & 49.3 &  \underline{65.2}\\


    \hline

    \multicolumn{1}{l|}{P-BEV(ours)} & 94.2 & 47.2 & 58.8 & 59.6 & 40.9 & 66.1 &  75.1 &  22.3 & 95.3 & 86.4 & 82.8 & 0.6 & 89.7 & 48.1 & 84.8 & 65.5 & 71.3 & 62.9 & 48.4 & 61.2\\  

    \multicolumn{1}{l|}{C-BEV(ours)} & 92.6 & 27.9 & 54.8 & 46.9 & 67.5 & 21.1 &  86.3 &  0.0 & 94.0 & 50.0 & 81.4 & 0.0 & 89.0 & 49.8 & 88.2 & 58.8 & 76.9 & 58.7 & 40.9 & 56.5 \\  

    \multicolumn{1}{l|}{PC-BEV(ours)} & 94.3 & 67.6 & 71.4 & 75.4 & 67.5 & 77.4 &  81.3&  0.1 & 95.2 & 39.7 & 83.0 & 0.1 & 86.3 & 49.7 & 85.1 & 70.2 & 68.2 & 70.2 & 64.2 & \textbf{66.4}\\

    \hline
    \end{tabular}}
  \label{tab:kittival}
\end{table*}

\begin{table*}[t]
  \centering
  \small
  \caption{Quantitative comparison of our PC-BEV. The results are reported in terms of the mIoU on the SemanticKITTI test set. The methods are grouped into point-based, projection-based and multi-view fusion models.}
    \setlength{\tabcolsep}{1.mm}{
    \begin{tabular}{l|lllllllllllllllllll|l}
    \hline
    & & & & & & & & & & & & & & & & & & & &\\
    & & & & & & & & & & & & & & & & & & & &\\
    \multicolumn{1}{l|}{Methods}  & {\rotatebox{70}{car}} & {\rotatebox{70}{bicycle\hspace{-1cm}}} & {\rotatebox{70}{motorcycle\hspace{-1cm}}} & {\rotatebox{70}{truck\hspace{-1cm}}} & {\rotatebox{70}{other-vehicle\hspace{-1cm}}} & {\rotatebox{70}{person\hspace{-1cm}}} & {\rotatebox{70}{bicyclist\hspace{-1cm}}} & {\rotatebox{70}{motorcyclist\hspace{-1cm}}} & {\rotatebox{70}{road}} & \multicolumn{1}{l}{\rotatebox{70}{parking}} & {\rotatebox{70}{sidewalk}} & {\rotatebox{70}{other-ground\hspace{-1cm}}} & {\rotatebox{70}{building\hspace{-1cm}}} & {\rotatebox{70}{fence\hspace{-1cm}}} & {\rotatebox{70}{vegetation\hspace{-1cm}}} & {\rotatebox{70}{trunk\hspace{-1cm}}} & {\rotatebox{70}{terrain\hspace{-1cm}}} & {\rotatebox{70}{pole\hspace{-1cm}}} & {\rotatebox{70}{traffic-sign\hspace{-1cm}}} & \multicolumn{1}{l}  {mIoU} \\    \hline
    \multicolumn{1}{l|}{LatticeNet } &   88.6 & 12.0 & 20.8 & 43.3 & 24.8 & 34.2 & 39.9 & 60.9 & 88.8 & 64.6 & 73.8 & 25.5 & 86.9 & 55.2 & 76.4 & 67.9 & 54.7 & 41.5 & 42.7 & 51.3 \\
    \multicolumn{1}{l|}{PointNL } &    92.1 & 42.6 & 37.4 & 9.8 & 20.0 & 49.2 & 57.8 & 28.3 & 90.5 & 48.3 & 72.5 & 19.0 & 81.6 & 50.2 & 78.5 & 54.5 & 62.7 & 41.7 & 55.8 & 52.2\\

    \multicolumn{1}{l|}{RandLa-Net } &  94.2 & 26.0 & 25.8 & 40.1 & 38.9 & 49.2 & 48.2 & 7.2 & 90.7 & 60.3 & 73.7 & 20.4 & 86.9 & 56.3 & 81.4 & 61.3 & 66.8 & 49.2 & 47.7 & 53.9 \\
    
    \multicolumn{1}{l|}{KPConv } &    96.0 & 30.2 & 42.5 & 33.4 & 44.3 & 61.5 & 61.6 & 11.8 & 88.8 & 61.3 & 72.7 & 31.6 & 90.5 & 64.2 & 84.8 & 69.2 & 69.1 & 56.4 & 47.4 & 58.8\\
    \hline

    
    \multicolumn{1}{l|}{MiniNet-KNN } &  90.5 & 42.3 & 42.1 & 28.5 & 29.4 & 47.8 & 44.1 & 14.5 & 91.6 & 64.2 & 74.5 & 25.4 & 89.4 & 60.8 & 82.8 & 60.8 & 66.7 & 48.0 & 56.6   & 55.8 \\
       
    \multicolumn{1}{l|}{SalsaNext } &     91.9 & 48.3 & 38.6 & 38.9 & 31.9 & 60.2 & 59.0 & 19.4 & 91.7 & 63.7 & 75.8 & 29.1 & 90.2 & 64.2 & 81.8 & 63.6 & 66.5 & 54.3 & 62.1 & 59.5\\

    \multicolumn{1}{l|}{PolarNet } &  93.8 & 40.3 & 30.1 & 22.9 & 28.5 & 43.2 & 40.2 & 5.6 & 90.8 & 61.7 & 74.4 & 21.7 & 90.0 & 61.3 & 84.0 & 65.5 & 67.8 & 51.8 & 57.5  & 54.3 \\  
    
     \multicolumn{1}{l|}{KPRNet } &  95.5  & 54.1  & 47.9  & 23.6  & 42.6  & 65.9  & 65.0  & 16.5  & 93.2  & 73.9  & 80.6  & 30.2  & 91.7  & 68.4  & 85.7  & 69.8  & 71.2  & 58.7  & 64.1 & 63.1\\

     \multicolumn{1}{l|}{RangeViT } &  95.4 & 55.8 & 43.5 & 29.8 & 42.1 & 63.9 & 58.2 & 38.1 & 93.1 & 70.2 & 80.0 & 32.5 & 92.0 & 69.0 & 85.3 & 70.6 & 71.2 & 60.8 & 64.7 & 64.0\\

    \multicolumn{1}{l|}{CENet } & 91.9 & 58.6 & 50.3 & 40.6 & 42.3 & 68.9 & 65.9 & 43.5 & 90.3 & 60.9 & 75.1 & 31.5 & 91.0 & 66.2 & 84.5 & 69.7 & 70.0 & 61.5 & 67.6 & 64.7\\

    \multicolumn{1}{l|}{MaskRange }  & 94.2 & 56.0 & 55.7 & 59.2 & 52.4 & 67.6 & 64.8 & 31.8 & 91.7 & 70.7 & 77.1 & 29.5 & 90.6 & 65.2 & 84.6 & 68.5 & 69.2 & 60.2 & 66.6 & 66.1\\
    
    \hline

    
    
    

    \multicolumn{1}{l|}{GFNet } &  96.0 & 53.2 & 48.3 & 31.7 & 47.3 & 62.8 & 57.3 & 44.7 & 93.6 & 72.5 & 80.8 & 31.2 & 94.0 & 73.9 & 85.2 & 71.1 & 69.3 & 61.8 & 68.0 & 65.4 \\

    \multicolumn{1}{l|}{AMVNet } &  96.2 & 59.9 & 54.2 & 48.8 & 45.7 & 71.0 & 65.7 & 11.0 & 90.1 & 71.0 & 75.8 & 32.4 & 92.4 & 69.1 & 85.6 & 71.7 & 69.6 & 62.7 & 67.2 & 65.3 \\


    \multicolumn{1}{l|}{CPGNet } &  96.7 & 62.9 & 61.1 & 56.7 & 55.3 & 72.1 & 73.9 & 27.9 & 92.9 & 68.0 & 78.1 & 24.6 & 92.7 & 71.1 & 84.6 & 72.9 & 70.2 & 64.5 & 71.9 & \textbf{68.3} \\

    \hline

    \multicolumn{1}{l|}{PC-BEV(ours)} & 96.5 & 66.5 & 57.7 & 37.6 & 48.2 & 62.5 &  68.0 &  49.1 & 93.0 & 73.5 & 79.8 & 32.1 & 92.1 & 69.2 & 85.4 & 70.0 & 70.0 & 62.4 & 63.0 & \underline{67.2}\\    
    \hline
    \end{tabular}}
  \label{tab:kittitest}
\end{table*}

\subsection{Transformer-CNN Mixture Architecture}
\label{sec:tcma}

Inspired by \cite{fan2023segtransconv,xu2023lightweight,cheng2023transrvnet}, we propose a Transformer-CNN hybrid network for feature extraction in BEV representation. We first capture the global scene information using the Transformer's self-attention mechanism, followed by further feature extraction through a lightweight CNN.

We illustrate the detailed feature extraction process using the Cartesian BEV features $F_{cart}^{in}$ as an example. We first divide the features into $n \times n$ patches, $Patch=\{patch_0, patch_1, \ldots, patch_{n \times n - 1}\}$, where $patch_i \in \mathbb{R}^{\frac{H}{n} \times \frac{W}{n} \times C} $. Each patch is then encoded into a vector using convolution operation with kernel size of $\frac{H}{n} \times \frac{W}{n}$. We denote the encoded patches as $Patch_{emb} = \{patch^0_{emb}, patch^1_{emb}, \ldots, patch^{n \times n -1}_{emb}\}$, where $patch^i_{emb} \in \mathbb{R}^{1 \times C_{emb}}$. Since the attention mechanism lacks the ability to distinguish positional information within the input sequence, we introduce the sine positional encoding $PE$ into the features. The final patch embedding input for self-attention can be expressed as: $Patch_{in} = \operatorname{MLP}(\operatorname{Cat}(Patch_{emb}, PE)$.
Then, we adopt the multi-head self-attention, the output is passed through a Feed-Forward Network (FFN) module. We denote the final patch embeddings from the transformer block as $ Patch_{out} = \{patch^{0}_{out}, patch^{1}_{out}, \ldots, patch^{n\times n -1}_{out}\}$. We reshape the output features from a 2D shape of $Patch_{out} \in \mathbb{R}^{n^2 \times C}$ to a standard 3D feature map $\mathbb{R}^{n\times n \times C}$. Afterward, we upsample the output bilinearly to match the full resolution of the projected pseudo-images:
\begin{equation}
\begin{aligned}
Patch^{out}_{bev} = \operatorname{Bi-Interpolate}(Patch_{out}) 
\end{aligned}
\label{equ:wg}
\end{equation}
where $Patch^{out}_{cart} \in \mathbb{R}^{H \times W \times C}$. We employ a straightforward addition operation to fuse $Patch^{out}_{cart}$ and $F^{in}_{cart}$:
\begin{equation}
\begin{aligned}
F^{enhanced}_{bev} = F^{in}_{cart} + Patch^{out}_{cart} \end{aligned}
\label{equ:wg}
\end{equation}

The feature enriched with global information is then fed into an efficient CNN model for further extraction. We utilize a U-net architecture CNN as described in \cite{li2022cpgnet}. Experiments demonstrate that our Transformer-CNN Mixture Architecture offers advantages in both performance and inference speed.

\begin{table*}[t]
  \centering
  \small
  \caption{Quantitative comparison of our PC-BEV. The results are reported in terms of the mIoU on the nuScenes validation set. }
    \setlength{\tabcolsep}{1.5mm}{
    \begin{tabular}{l|llllllllllllllll|l}
    \hline
    & & & & & & & & & & & & & & & & &\\
    & & & & & & & & & & & & & & & & &\\
    \multicolumn{1}{l|}{Methods}  & {\rotatebox{70}{barrier}} & {\rotatebox{70}{bicycle\hspace{-1cm}}} & {\rotatebox{70}{bus\hspace{-1cm}}} & 
    {\rotatebox{70}{car\hspace{-1cm}}} & {\rotatebox{70}{construction\hspace{-1cm}}} & {\rotatebox{70}{motorcycle\hspace{-1cm}}} & 
    {\rotatebox{70}{pedestrian\hspace{-1cm}}} 
    & {\rotatebox{70}{traffic-cone\hspace{-1cm}}} & 
    {\rotatebox{70}{trailer}} &
    {\rotatebox{70}{truck}} & 
    {\rotatebox{70}{driveable\hspace{-1cm}}} & {\rotatebox{70}{other\hspace{-1cm}}} & {\rotatebox{70}{sidewalk\hspace{-1cm}}} & {\rotatebox{70}{terrain\hspace{-1cm}}} & {\rotatebox{70}{manmade\hspace{-1cm}}} & {\rotatebox{70}{vegetation\hspace{-1cm}}} & 
    \multicolumn{1}{l}  {mIoU} \\    \hline

    \multicolumn{1}{l|}{RangeNet++ } &  66.0 & 21.3 & 77.2 & 80.9 & 30.2 & 66.8 & 69.6 & 52.1 & 54.2 & 72.3 & 94.1 & 66.6 & 63.5 & 70.1 & 83.1 & 79.8  & 65.5\\  
    \multicolumn{1}{l|}{Salsanext} & 74.8 & 34.1 & 85.9 & 88.4 & 42.2 & 72.4 & 72.2 & 63.1 & 61.3 & 76.5 & 96.0 & 70.8 & 71.2 & 71.5 & 86.7 & 84.4 & 72.2 \\ 

    \multicolumn{1}{l|}{PolarNet} & 74.7 & 28.2 & 85.3 & 90.9 & 35.1 & 77.5 & 71.3 & 58.8 & 57.4 & 76.1 & 96.5 & 71.1 & 74.7 & 74.0 & 87.3 & 85.7 & 71.0 \\ 

    \multicolumn{1}{l|}{PCSCNet} & 73.3 & 42.2 & 87.8 & 86.1 & 44.9 & 82.2 & 76.1 & 62.9 & 49.3 & 77.3 & 95.2 & 66.9 & 69.5 & 72.3 & 83.7 & 82.5 & 72.0 \\ 

    \multicolumn{1}{l|}{SVASeg} & 73.1 & 44.5 & 88.4 & 86.6 & 48.2 & 80.5 & 77.7 & 65.6 & 57.5 & 82.1 & 96.5 & 70.5 & 74.7 & 74.6 & 87.3 & 86.9 & 74.7 \\ 

    \multicolumn{1}{l|}{AMVNet} & 79.8 & 32.4 & 82.2 & 86.4 & 62.5 & 81.9 & 75.3 & 72.3 & 83.5 & 65.1 & 97.4 & 67.0 & 78.8 & 74.6 & 90.8 & 87.9 & 76.1 \\ 

    
    \multicolumn{1}{l|}{RPVNet} & 78.2 & 43.4 & 92.7 & 93.2 & 49.0 & 85.7 & 80.5 & 66.0 & 66.9 & 84.0 & 96.9 & 73.5 & 75.9 & 70.6 & 90.6 & 88.9 & \underline{77.6} \\ 
    
    \hline
    

    \multicolumn{1}{l|}{P-BEV(ours)} & 77.2 & 40.5 & 89.9 & 90.3 & 48.9 & 84.8 &  80.7 &  64.7 & 64.4 & 83.4 & 96.9 & 70.6 & 76.3 & 74.8 & 81.4 & 76.8  & 75.0\\    

    \multicolumn{1}{l|}{C-BEV(ours)} & 73.2 & 18.8 & 89.4 & 89.7 & 45.2 & 74.7 &  70.8 &  52.6 & 60.6 & 82.3 & 96.2 & 72.7 & 73.6 & 74.0 & 85.7 & 83.2  & 71.4\\    

    \multicolumn{1}{l|}{PC-BEV(ours)} & 78.2 & 46.3 & 92.5 & 93.4 & 55.0 & 87.1 &  81.0 &  65.4 & 69.2 & 85.7 & 97.1 & 76.8 & 77.0 & 76.3 & 90.6 & 88.5  & \textbf{78.8}\\

    \hline
    \end{tabular}}
  \label{tab:nuScenes}
\end{table*}

\section{Experiments}


\noindent \textbf{Dataset} We evaluate our proposed dual-path network over two LiDAR datasets of driving scenes that have been widely adopted for benchmarking in semantic segmentation. SemanticKITTI \cite{behley2019semantickitti} is based on the KITTI Vision Benchmark, and it has a total of 43,551 scans sampled from 22 sequences collected in different cities in Germany. nuScenes-lidarseg \cite{caesar2020nuscenes} has 40,000 scans captured in a total of 1000 scenes of 20s duration. It is collected with a 32 beams LiDAR sensor and is sampled at 20Hz. We follow the official guidance to leverage mean intersection-over-union (mIoU) as the evaluation metric as defined in \cite{behley2019semantickitti}.

\begin{table}
\centering
\caption{Illustration of the efficiency of our remap-based interaction. The experiment is conducted on the BEV features with voxel resolution set to 0.2m. The latency is tested on a 2080Ti GPU. SC denotes the scatter back operation. $*$ denotes the method implemented in \cite{li2022cpgnet}, $\dagger$ denotes the method implemented in \cite{pytorchscatter2020}.}
\setlength{\tabcolsep}{1.mm}{
\begin{tabular}{l|c|c|c|c}
\hline
\multirow{2}{*}{Method} & \multicolumn{3}{c|}{Point-based} & \multirow{2}{*}{ \makecell{Remap-based \\ (Ours)} } \\

\cline { 2 - 4 } & \makebox[0.05\textwidth][c]{GS} & \makebox[0.05\textwidth][c]{SC}  & Overall \\
\hline

\multirow{2}{*}{latency \(ms\)} & 0.3 & 6.5$^{*}$ & 6.8 & \multirow{2}{*}{0.04} \\

\cline { 2 - 4 } & 0.3 & 7.7$^{\dagger}$ & 8.0 & \\
\hline
\end{tabular}}
\label{tab:inference_speed}
\end{table}

\subsection{Main Results}
In Table \ref{tab:kittival}, we report the experiment results on the SemanticKITTI val set. For the Polar and Cartesian partitioning strategies alone, we achieve performance of 61.2\% and 56.5\%, respectively. It can be seen that with two partitioning fusions, we achieve a large margin improvement over using them independently. Compared with outer methods, we outperform the Range-BEV fusion GFNet \cite{qiu2022gfnet} by 3.4\%mIoU, and suppress the Range-BEV fusion AMVNet by 1.2\% mIoU.

In Table \ref{tab:kittitest}, we show the results on the test set of SemanticKITTI. We compare with methods including point-based method LatticeNet\cite{rosu2019latticenet}, PointNL\cite{cheng2020cascaded}, RandLa-Net\cite{hu2020randla}, KPConv\cite{thomas2019kpconv}, Projection-base method SalsaNext\cite{cortinhal2020salsanext}, RangeViT\cite{ando2023rangevit}, CENet\cite{cheng2022cenet} ,MaskRange\cite{gu2022maskrange}, and fusion-base method GFNet\cite{qiu2022gfnet}, AMVNet\cite{liong2020amvnet}, RPVNet\cite{xu2021rpvnet}, CPGNet\cite{li2022cpgnet}. It is evident that we have a clear advantage over single-view projection-based methods. Specifically, we outperform RangeViT by 3.2\%mIoU, suppress CENet by 2.5\%mIoU, and show better performance than Maskrange by 1.1\%mIoU. For the multiview fusion based methods, our method outperforms GFNet and AMVNet, surpassing them by 1.8\% and 1.9\%, respectively.

In Table \ref{tab:nuScenes}, we report the results of our proposed dual-path network on the nuScenes validation dataset. We compare with methods including PCSCNet\cite{park2023pcscnet}, SVASeg\cite{zhao2022svaseg}, etc. For the Polar and Cartesian partitioning strategies alone, we achieve performance of 75.0\%mIoU and 71.4\%mIoU, respectively. A significant performance improvement can also be achieved by fusing the two partitioning strategies. Specifically, it can be seen that compared with the previous methods, our method shows an obvious performance improvement. Specifically, we suppress the Range-BEV fusion network (AMVNet) by 2.7\%mIoU. Surprisingly, we even suppress the Range-Point-Voxel Fusion Network (RPVNet) by 2.2\%mIoU.

We compared the latency of different operations for feature interaction, and the experimental results are shown in Table \ref{tab:inference_speed}. For the grid sampling operation, we use the PyTorch's built-in function $\operatorname{grid\_sample(\cdot)}$, and for the scatter operation, we utilized the efficient implementation in CPGNet \cite{li2022cpgnet}. It can be observed that the scatter back operation consumes a significant amount of time. This is because, unlike grid sampling, which only involves read operations, scatter back includes write operations that require locking due to potential conflicts. Our method also demonstrates a noticeable performance improvement compared to grid sampling. This is because our approach better leverages the spatial continuity of features, resulting in a higher cache hit rate and thus faster processing speed.

\subsection{Ablation Studies}

We perform ablation studies on the nuScenes validation set to examine the efficacy of each component in our framework. The results are shown in Table \ref{tab:Ablation}. The Cartesian or Polar indicates that we only use one branch, and the third row (both Cartesian and Polar) means we simply calculate the mean of the outputs from the two branches to obtain the final predictions. It can be seen that only the native two-branch network can improve the performance by 1.2\% mIoU, which demonstrates the great potential of the Polar-Cartesian fusion. and we can boost the performance by 0.9\% mIoU with our remap-based fusion module. Together with the Transformer-CNN mixture architecture, the performance can be further improved by another 0.6\% mIoU. For the inference speed of the model, we can see that for the Polar and Cartesian partitioning strategies individually, the inference speed is 20ms and 19ms, respectively. Our PC-BEV fusion framework achieved a final latency of 31ms, which is 20\% faster than the point-based (the fourth row) method. The latency is tested on an NVIDIA V100 GPU.

\begin{table}[t]   
\small
    \centering
    \caption{Ablation study of our Polar-Cartesian BEV Fusion Framework on the nuScenes validation set. PB denotes the point-based fusion, RM denotes the remap fusion, T-C denotes the Transformer-CNN Mixture Architecture.}  
    \setlength{\tabcolsep}{1.5mm}{
\begin{tabular}{c|c|c|c|c|c|c}
\hline Cartesian & Polar & PB & RM & T-C & \makebox[0.06\textwidth][c]{mIoU(\%)} & latency (ms) \\

\hline $\sqrt{ }$ &  &  &  & &   $ 71.4 $ & 20 \\ 
 & $\sqrt{ }$ &  &  & &   $ 75.0 $ & 19 \\ 
$\sqrt{ }$& $\sqrt{ }$  &  &  &  &   $ 76.2 $ &  27\\
$\sqrt{ }$& $\sqrt{ }$  & $\sqrt{ }$ &  &  & $77.3 $ & 39 \\
$\sqrt{ }$& $\sqrt{ }$  &  & $\sqrt{ }$ &  & $78.2 $& 29 \\
$\sqrt{ }$& $\sqrt{ }$  &  & $\sqrt{ }$ & $\sqrt{ }$  & $78.8$ & 31 \\
\hline
\end{tabular}}
\label{tab:Ablation}
\end{table}

\section{Conclusion}

In conclusion, we have challenged the necessity of multiview fusion for LiDAR segmentation by proposing a real-time approach based on fusing features from distinct partitioning methods (Polar and Cartesian) within a unified BEV space. Our efficient remap-based spatial alignment fusion method, leveraging memory continuity, significantly outperforms previous point-based interaction methods in terms of speed while preserving richer contextual information. Furthermore, our proposed Transformer-CNN hybrid architecture enhances model performance without compromising real-time processing capabilities. Extensive experiments on SemanticKITTI and nuScenes datasets validate the effectiveness and efficiency of our approach. Future research may explore the application of this method to BEV representations derived from multi-camera image data.

\section{Acknowledgments}
This work was partially supported by the Computing for the Future at Fudan (CFFF).

\bibliography{aaai25}

\end{document}